\documentclass[default]{sn-jnl}
\jyear{2021}%

\theoremstyle{thmstyleone}%
\usepackage{natbib}
\theoremstyle{thmstyletwo}%

\theoremstyle{thmstylethree}%

\begin{document}

\title[R\MakeLowercase{es}VGAE: Going Deeper with Residual Modules for Link Prediction]{R\MakeLowercase{es}VGAE: Going Deeper with Residual Modules for Link Prediction}


\author*[1]{\fnm{Indrit} \sur{Nallbani}}\email{nallbani17@itu.edu.tr}

\author[1]{\fnm{Reyhan Kevser } \sur{Keser}}\email{keserr@itu.edu.tr}

\author[2]{\fnm{Aydin } \sur{Ayanzadeh}}\email{aydina1@umbc.edu}
\author[3]{\fnm{Nurullah  } \sur{Çalık}}\email{nurullah.calik@medeniyet.edu.tr}

\author[1]{\fnm{Behçet Uğur} \sur{Töreyin}}\email{toreyin@itu.edu.tr}

\affil*[1]{\orgdiv{Institute of Informatics}, \orgname{Istanbul Technical University}, \orgaddress{\street{Katar Caddesi}, \city{Istanbul}, \postcode{34467}, \state{Istanbul}, \country{Turkey}}}

\affil[2]{\orgdiv{Department of Computer Science}, \orgname{University of Maryland Baltimore County}, \orgaddress{\street{1000 Hilltop Cir}, \city{Baltimore}, \postcode{21250}, \state{Maryland}, \country{USA}}}

\affil[3]{\orgdiv{Department of Biomedical sciences}, \orgname{Istanbul Medeniyet University}, \orgaddress{\street{Dumlupınar D100 Karayolu}, \city{Istanbul}, \postcode{34720}, \state{Istanbul}, \country{Turkey}}}


\abstract{ Graph autoencoders are efficient at embedding graph-based data sets. Most graph autoencoder architectures have shallow depths which limits their ability to capture meaningful relations between nodes separated by multi-hops. In this paper, we propose Residual Variational Graph Autoencoder, ResVGAE, a deep variational graph autoencoder model with multiple residual modules. We show that our multiple residual modules, a convolutional layer with residual connection, improve the average precision of the graph autoencoders. Experimental results suggest that our proposed model with residual modules outperforms the models without residual modules and achieves similar results when compared with other state-of-the-art methods.}
\keywords{Residual Variational Graph Autoencoders, Graph Embedding, Residual Connections, Graph Convolutional Layers, Link Prediction}

\maketitle

\section{Introduction}
\label{sec1}
Learning-based feature extraction approaches have led to better performance in machine learning tasks, such as computer vision, machine translation, and object detection. Most real-world data sets have proven to be very successful in producing data representations that are successfully used in several tasks, such as fraud detection \cite{frauddetection}, recommendation systems~\cite{recommendationsystem}, churn prediction \cite{churnprediction} and predicting earthquakes using graph processes \cite{ruiz2019gated}. Graph neural networks (GNN) can efficiently exploit the relationship between data set instances in non-Euclidean space.
Different variants of graph autoencoders, \cite{triadic2020}, \cite{yu2021influence}, ~\cite{weng2020adversarial}, \cite{davidson2018hyperspherical}, \cite{deepwalk2014}, have been very successful in capturing meaningful representations for node classification \cite{grover2016node2vec}, link prediction~\cite{vilnis2014guassianembedding} and graph classification \cite{graph-structure-2017} tasks.

 \begin{figure}[hbt!]

  \centering

	\includegraphics[width = 1\linewidth]{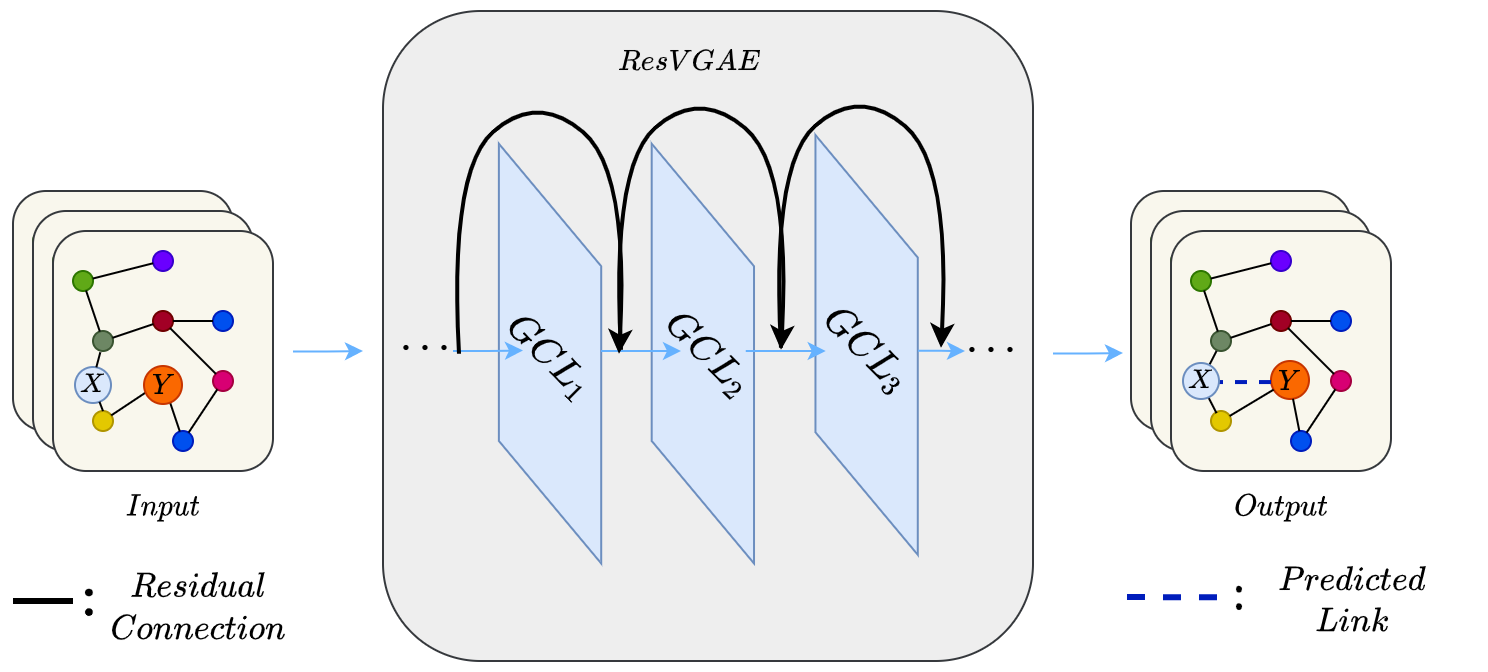}

	\caption{End-to-end framework of ResVGAE.
	Given two distant nodes $X$ and $Y$, the model predicts a link (blue dashed line in the output) between them. This is made possible efficiently, thanks to the residual modules (residual connections with their consecutive Graph Convolutional Layers (GCLs). Here, we show a model with three modules, namely, $GCL_1$, $GCL_2$, and $GCL_3$.}
	\label{fig:draw1}
\end{figure}

In recent years, we see a proliferation of graph embedding techniques for improving graph convolutional networks and their applications in graph-structured data sets \cite{8553465}. 
Perozzi \textit{et al.} propose DeepWalk, a framework that embeds graph nodes based on the information acquired from truncated random walks. The framework is robust and can learn meaningful node representations for large-scale data sets \cite{perozzi2014deepwalk}, whereas Grover \textit{et al.} presents a framework that maps node features in a low-dimensional space while preserving the networks neighborhoods of nodes \cite{grover2016node2vec}. Tang \textit{et al.} propose another classical approach that is capable of network embedding while preserving first and second-order proximities of the nodes \cite{line}.
Another method that embeds graph data sets is Deep Variational Network Embedding in Wasserstein Space (DVNE), a network embedding framework that uses a 2-Wasserstein distance as a similarity measure between the latent node distributions. The proposed framework preserves the first-order and second-order node proximities in the network \cite{rw_2}.
 GraphSAGE is a framework that generates inductive node embeddings for previously unseen data by leveraging node features. In this work, the authors train a set of aggregator functions that learn from the neighbor's node features \cite{hamilton2017inductive}, whereas Pan \textit{et al.} propose an adversarially regularized variational graph autoencoder (ARVGA), a framework where the latent representations are enforced to match prior distributions using an adversarial training scheme \cite{pan2018adversarially}.
Several theoretical works have pointed out the occurrence of the over-smoothing phenomenon, the process in which node representations become indistinguishable when nodes interact with other nodes they are not directly connected,  \cite{nt2019revisiting}, \cite{alon2020bottleneck}, \cite{mavromatis2020graph}, \cite{zhou2020graph}. This limits the ability of the models to stack many GNN layers. We want deep graph networks to capture long-distance interactions between far nodes for link prediction tasks.
As shown in Fig.\ref{fig:draw1}, to capture meaningful information between distant nodes $X$ and node $Y$, our model embeds the nodes into a similar low-dimensional vector space. If the node embeddings are similar, then our model predicts a link between them. In this example, nodes $X$ and $Y$ are three hops apart so ResVGAE has three residual modules.

\begin{figure*}[hbt!]
	\centering
	\includegraphics[width = 1\linewidth]{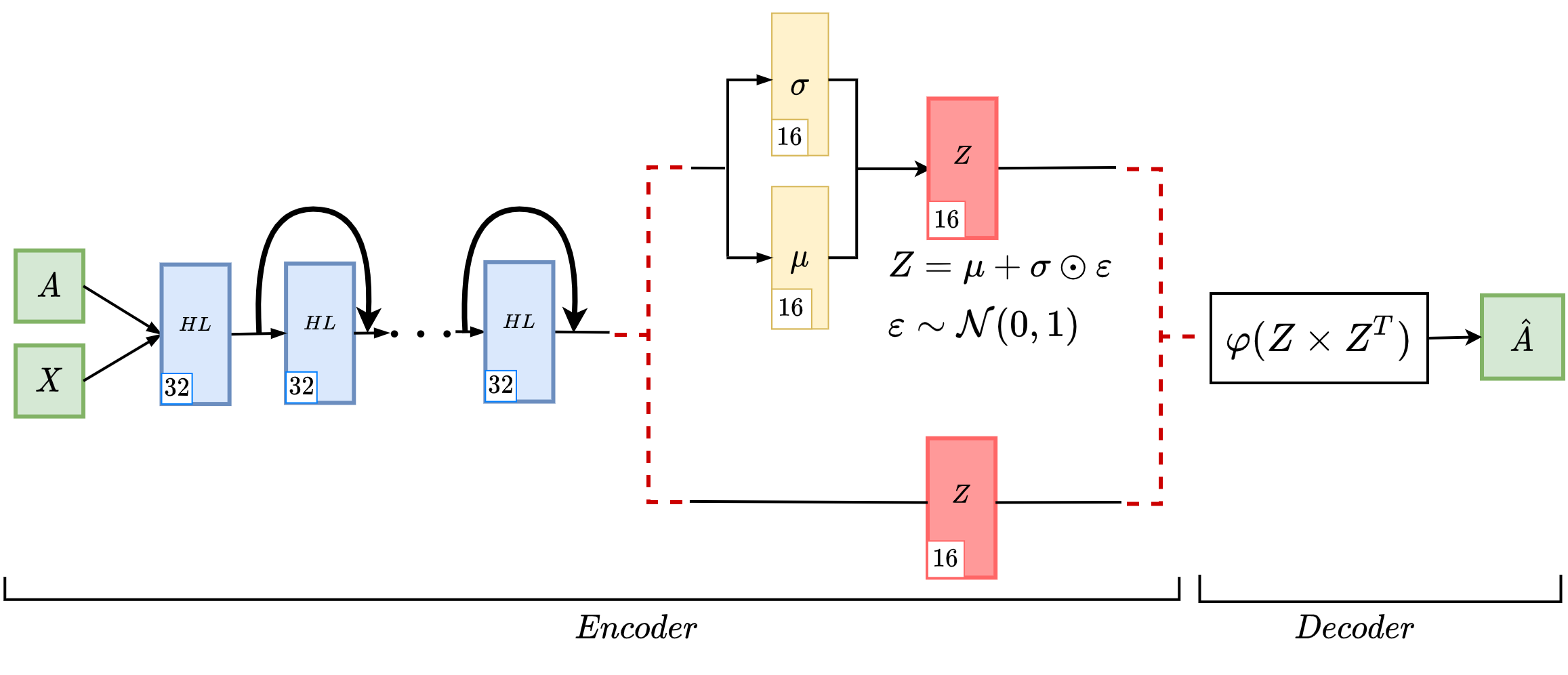}
	\caption{Model architecture of ResVGAE. Residual connections start after the first Hidden Layer ($HL$) since the input and the output size of layers with residual modules must be the same. The encoder takes the adjacency matrix $A$ and the feature matrix $X$ as inputs and outputs the node embeddings $Z$. The decoder takes as input the embedding matrix $Z$ and outputs the reconstructed adjacency matrix $\hat A$. 
	The blocks in blue indicate the graph convolutional layers that embed 32-dimensional node feature vectors into a matrix. Similarly, yellow blocks constitute the graph convolutional layers that embed 16-dimensional hidden layer features into the output Z matrix.
	The upper and lower branches of the encoder represent variational graph autoencoder and graph autoencoder architectures, respectively.}

	\label{fig:res}
\end{figure*}

We propose a model architecture with residual modules that combines residual connections  \cite{he2016deep} and variational graph autoencoders (cf. Fig.\ref{fig:res}) in order to alleviate the over-smoothing phenomenon.

The overall contributions of our study are as follows:
\begin{itemize}
\item  We propose a graph variational autoencoder model with residual modules and compare it with the other state-of-the-art models for link prediction tasks.
\item We measure the accuracy of adding residual modules on similar graph-based autoencoders with different depths.
\item We study the over-smoothing phenomena by implementing our proposed architecture from one up to eight residual modules.
\end{itemize}

The rest of the paper is organized as follows: Section~\ref{sec:method} covers the proposed model architectures. In Section~\ref{sec:experiments}, comparative results of the proposed ResVGAE model are presented for the link prediction task on three benchmark data sets. We conclude the paper and propose new research paths for future work in Section~\ref{sec:conclusion}.

\section{Method}
\label{sec:method}

Graph autoencoders are deep learning frameworks whose inputs are instance features and adjacency matrices and whose output is the reconstructed adjacency matrix. These frameworks mainly consist of two components, namely, the encoder and the decoder. The encoder part transforms the input into a lower-dimensional embedding while the decoder part transforms the embedding into the reconstructed adjacency matrix. 
Let $G=(V, E)$ denote a graph $G$, where $V$ is the set with $N$ nodes, and $E$ is the set of edges. Moreover, let $v_{i} \in V$ denote a node and $e_{ij}=(v_{i},v_{j})$ denote an edge between two nodes $v_{i}$ and $v_{j}$.
We define the adjacency matrix $A$ as:
\begin{equation}
A:=\left\{\begin{array}{ll}
A_{i j}=1, & \text  e_{i j} \in E \\
A_{i j}=0, & \text e_{i j} \notin E \\
\end{array}\right.
\end{equation}
where $A$ is a symmetric matrix.
Let  $X$ be the feature matrix of the nodes.
For all the encoder architectures, we use the graph convolutional layer proposed by \cite{thomas2016semi}. The layer follows the propagation rule:
\begin{equation}
\label{eq:1}
H^{(l+1)}=\varphi\left(\tilde{D}^{-\frac{1}{2}} \tilde{A} \tilde{D}^{-\frac{1}{2}} H^{(l)} W^{(l)}\right)
\end{equation}
where $\varphi$ is the sigmoid activation function, $\tilde{D} $ is the degree matrix, $\tilde{A}$ is the normalized adjacency matrix with self-loops and $W^{(l)}$ is the layer's weight matrix. In a multi-layer model $H^{(0)} = X$ and $H^{(l)}$ is the feature map of the $l^{th}$ layer.

Variational graph encoders transform the graph into a lower-dimensional embedding, using graph convolutional layers and a sampling layer. These graph convolutional layers transform the graph into the desired lower-dimensional space and produces mean and standard deviation values using two layers. Then, the sampling layer takes these mean and deviation values to generate samples from the prior distribution. Hence the generated samples constitute the embedding of the graph. 

The final embeddings are encoded as a distribution over the latent space as:
\begin{equation}
q(Z \mid X, A)=\prod_{i=1}^{N} q\left(z_{i} \mid X, A\right)
\end{equation}
where

\begin{equation}
q\left(z_{i} \mid X, A\right)=\mathcal{N}\left(\mu_{i}, \operatorname{diag}\left(\sigma^{2}\right)\right)
\end{equation}
 Here, $z_{i}$ is the embedding vectors for node $i$; $\mu_{i}$ and $\sigma$ are the graph convolutional layer vector outputs of the encoder.

Using this structure provides a graph embedding with the desired distribution (cf. Fig.\ref{fig:res}).

For the graph autoencoder models, the decoder is a non-probabilistic model that reconstructs the adjacency matrix by computing the inner-product of the latent representations of  two-node embeddings :
\begin{equation}
\hat{A}=\varphi\left(Z Z^{T}\right)
\end{equation}
In all the models, $\varphi (\cdot) $ denotes the sigmoid activation function.

For the graph variational autoencoder models, the decoder is a probabilistic model that reconstructs the adjacency matrix by computing the probabilistic inner product of the latent representation of nodes:
\begin{equation}
p(A \mid Z)=\prod_{i=1}^{N} \prod_{j=1}^{N} p\left(A_{i j} \mid z_{i}, z_{j}\right)
\end{equation} where
\begin{equation}p\left(A_{i j} \mid z_{i}, z_{j}\right)=\varphi\left(z_{i}^{T} z_{j}\right)
\end{equation}
Vanilla models are trained  by minimizing reconstruction loss:
\begin{equation}
\mathcal{L}=\mathbb{E}_{q(Z \mid X, A)}[\log p(A \mid Z)]
\end{equation}
and variational models are optimized by maximizing the variational lower bound while minimizing reconstruction loss:

\begin{equation}
\mathcal{L}=\mathbb{E}_{q(Z \mid X, A)}[\log p(A \mid Z)]-\mathrm{KL}[q(Z \mid X, A) \| p(Z)]
\end{equation}

\noindent where $\mathrm{KL}[q(\cdot) \| p(\cdot)]$ is the Kullback-Leibler divergence between $q(\cdot)$ and $p(\cdot)$ .  The posterior distribution is approximated by a Gaussian distribution using the reparametrization trick \cite{Kingma2014}. 
Our proposed method, ResVGAE is a variational graph autoencoder with multiple residual modules. The decoder and loss are the same as variational graph autoencoders.
To improve the performance of the variational graph autoencoder, we propose utilizing residual modules in the encoder similar to \eqref{eq:1}. A residual module can be represented as:
\begin{equation}
H^{(l+1)}=\varphi\left(\tilde{D}^{-\frac{1}{2}} \tilde{A}  \tilde{D}^{-\frac{1}{2}} H^{(l)} W^{(l)}\right) + H^{(l)}
\end{equation}
The input and the output sizes of the hidden layers must be the same to add residual modules. The number of
residual modules can be determined depending on the depth of exploration among nodes for link prediction.
\section{Experimental Results}
\label{sec:experiments}

We evaluate the model's performance  on three benchmark data sets: Cora \cite{cora}, CiteSeer \cite{citeseer} and PubMed \cite{pubmed}.
Cora and CiteSeer are networks of computer science publications where the nodes represent the publications and the edges represent the citations. The PubMed data set is also a citation network that contains a set of articles related to diabetes disease.

We compare ResVGAE against three baseline algorithms: Deep Walk (DW)\cite{deepwalk2014}, Spectral Clustering (SC) \cite{tang2011leveraging} and Adversarially Regularized Graph Autoencoder (ARVGE) \cite{pan2018adversarially}

All models and data sets in this paper have been used for link prediction tasks and Table~\ref{tbl:summary_dataset} gives a detailed summary of the data sets we used.
\begin{table}[hbt!]
	\centering
	\caption{Data set statistics}
	\label{tbl:summary_dataset}
	\resizebox{1.\linewidth}{!}
	{
		\begin{tabular}{rrrrrr}
		\centering
			Data set &   \# Nodes & \# Edges &  \# Classes & \# Features \\
			\hline CitesSeer  & 3,327 & 4,732 & 6 & 3,703 \\
			Cora  & 2,708 & 5,429 & 7 & 1,433 \\
			PubMed  & 19,717 & 44,338 & 3 & 500
		\end{tabular}
	}
\end{table}

We evaluate all the models based on the same baseline as in \cite{kipf2016variational} and we use AP and AUC scores to report the average precision of 10 runs with random train, validation, and test splits of the same size, and all the models are trained for 200 epochs. The validation and test sets contain 5\% and 10\% of the total edges, respectively.
The embedding dimensions of the node features for the hidden layers is 32 and the embedding dimension of node features for the output layer is 16. The models are optimized using the Adam optimizer with a learning rate of 0.01.

In our experiments, we use average precision (AP) and area under ROC curve (AUC) metrics to measure the accuracy of our models. We use PyTorch Geometric \cite{pytorch-geometric} and our code will be available upon acceptance.

\begin{figure}[hbt!]
\centering
\includegraphics[width = 1\linewidth]{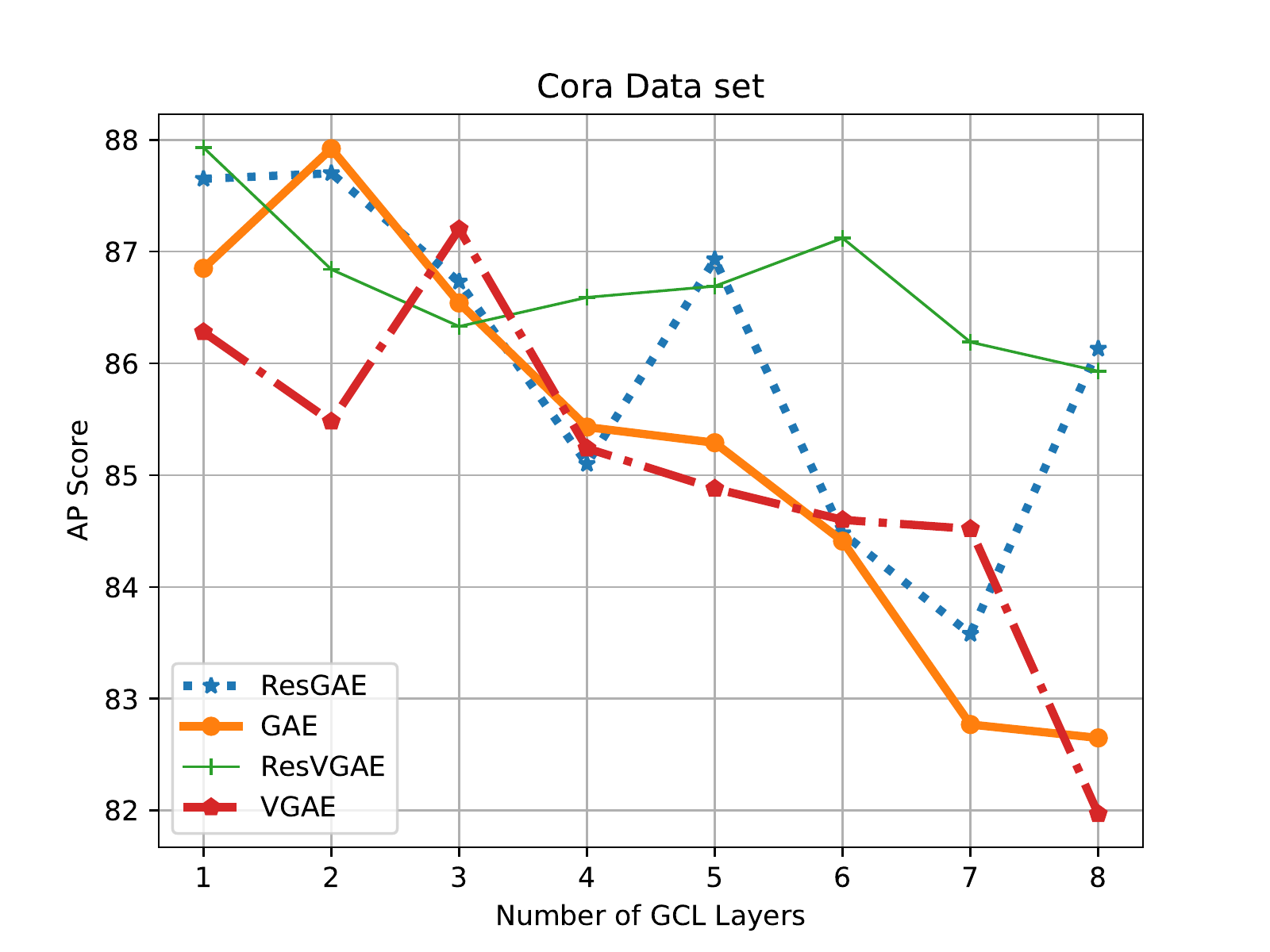}
\caption{The AP scores of the models with multiple numbers of layers for the Cora data set. ResGAE has the highest AP score followed by ResVGAE and the graph autoencoders without the residual modules. Residual modules substantially improve the performance of the variational autoencoder models in comparison to models without them. The performance gap increases for a higher number of layers.}
\label{fig:cora}
\end{figure}
 Experiments indicate that, for shallow models, all proposed models achieve similar average precision scores. Here we see that all models with one residual module embed the node features in a very similar way. The score differs when we use models with deeper networks.

 In models with eight graph convolutional layers, we see that models with residual modules have higher average precision scores than models without residual modules.
In Fig.~\ref{fig:cora}, we run all the models from one up to eight graph convolutional layers.
Here we see that as the models become deeper their average precision decreases significantly. The models with residual modules achieve higher scores than the models without residual modules.

\begin{figure}[hbt!]
\centering
\includegraphics[width = 1.1\linewidth]{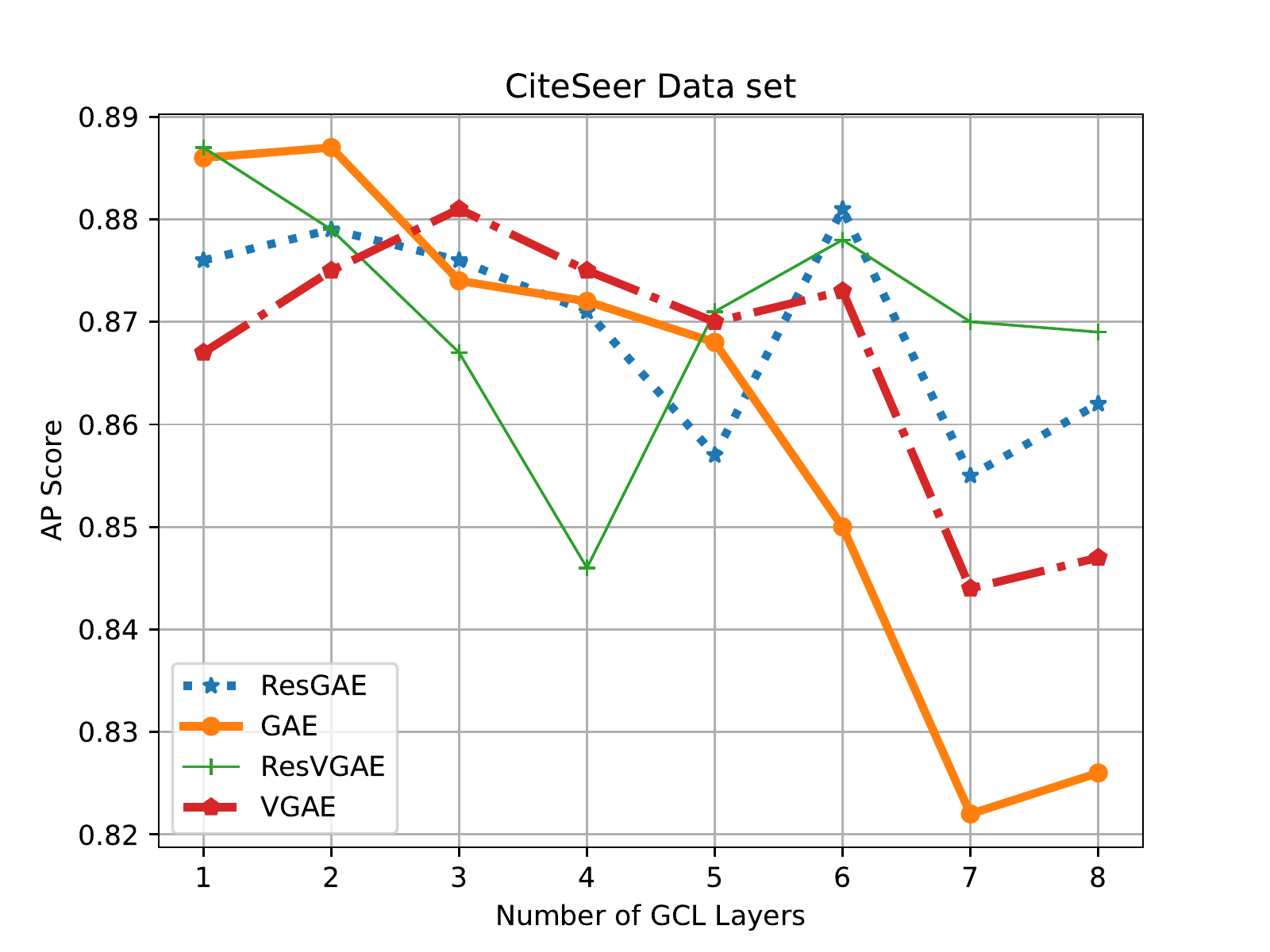}
\caption{The AP scores of the models with a different number of GCL layers for the CiteSeer data set. Models with residual modules score higher than the models without residual modules. The highest scoring model is ResVGAE followed by ResGAE for the number of layers seven and eight.}
\label{fig:citeseer}
\end{figure}

\begin{figure}[hbt!]
	\centering
	\includegraphics[width = 1.1\linewidth]{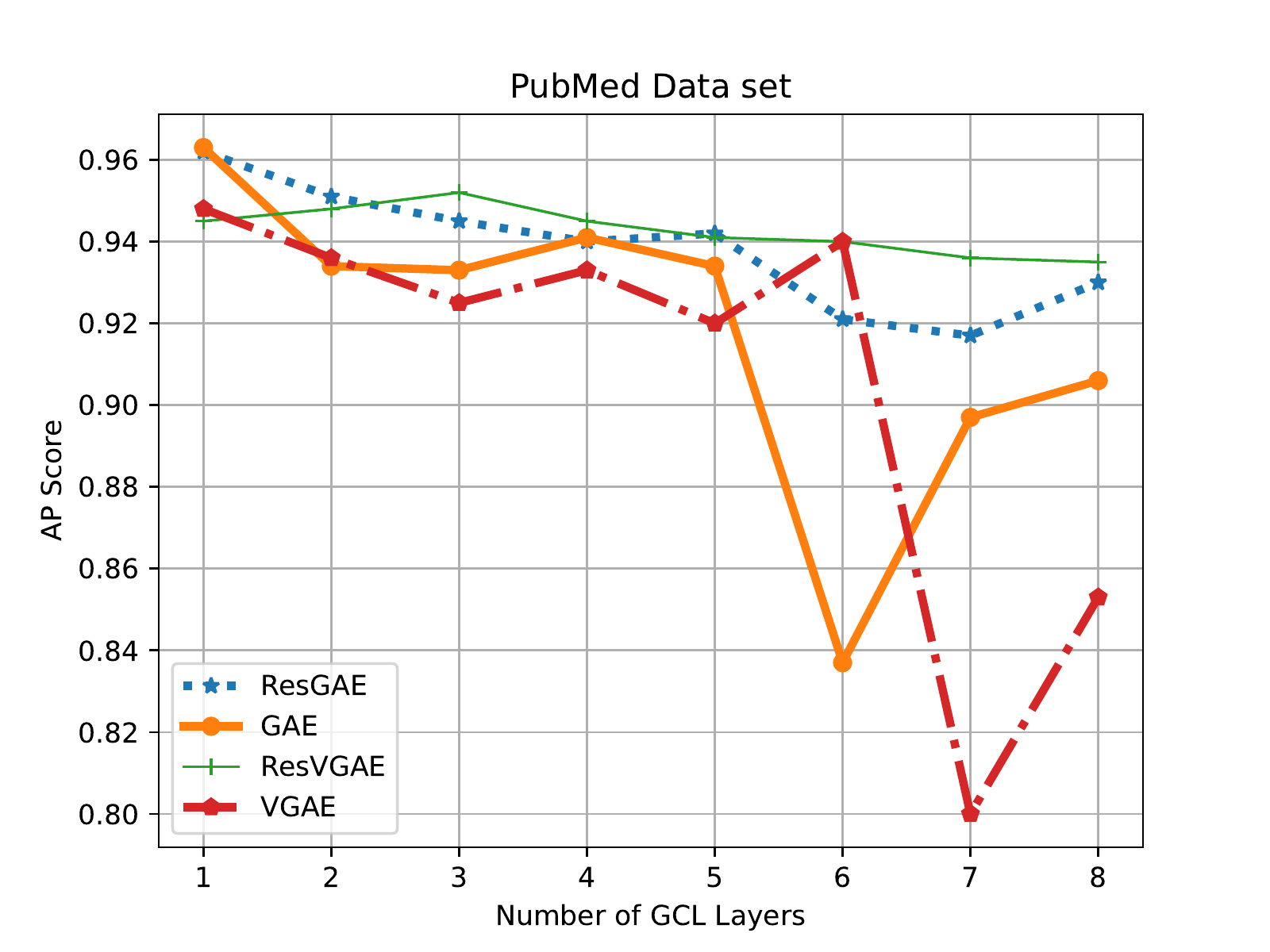}
	\caption{The AP scores of the models with a different number of GCL layers for the PubMed data set. Residual modules with more layers result in a higher AP score. }
	\label{fig:pubmed}
\end{figure}

Plots in Figs.~\ref{fig:citeseer} and ~\ref{fig:pubmed} indicate the steady drop in average precision values for all the models on CiteSeer and PubMed data sets, respectively. Moreover, for deeper models, the results suggest that deep models with residual modules outperform the models without residual modules.

\begin{table*}[hbt!]
\centering
\caption{Performance comparison of models with and without residual modules. Our proposed architectures, ResVGAE and ResGAE, achieve higher AUC and AP scores on Cora, CiteSeer, and PubMed data sets when compared to their variational graph autoencoder counterparts without residual modules. \\}
\label{tbl:comp_table4}
\resizebox{1\linewidth}{!}{
\begin{tabular}{lcccccc}
	\hline  { Method } & \multicolumn{2}{c} { Cora } & \multicolumn{2}{c} { CiteSeer } & \multicolumn{2}{c} { PubMed } \\
	& AUC & AP & AUC & AP & AUC & AP \\
	\hline
	ResVGAE (Ours)   & $\textbf{85.93} \pm \textbf{1.02}$ & $\textbf{88.55} \pm \textbf{0.65}$  & $\textbf{86.90}\pm \textbf{1.85}$ & $\textbf{80.30} \pm \textbf{0.40}$ & $\textbf{91.28} \pm \textbf{0.80}$ & $\textbf{93.50 }\pm \textbf{0.68}$ \\
	VGAE  		& $81.97 \pm  0.79$ & $85.51 \pm {2.54}$  & $83.70 \pm 0.77$ & $79.00 \pm 0.60$ & $85.30 \pm 0.95$ & $82.60 \pm 0.60$ \\

	\hline 
ResGAE (Ours) 	& $\textbf{86.13} \pm \textbf{0.844}$ & $\textbf{88.85} \pm \textbf{0.31}$  & $\textbf{87.10} \pm 1\textbf{.99}$ & $\textbf{84.40} \pm \textbf{1.95}$ & $\textbf{93.00} \pm \textbf{0.46}$ & $\textbf{91.40} \pm  \textbf{0.29}$ \\
	GAE         & $82.65 \pm 0.79$ & $86.46 \pm 0.32$  & $82.60  \pm 1.28$ & $81.60 \pm 1.17$ & $ 90.60 \pm 1.70$ & $87.80 \pm 0.90$ \\
	\hline
\end{tabular}
}
\end{table*}

Improvement in terms of average precision values is evident for models with residual models tested on Cora, CiteSeer, and PubMed data sets (Table \ref{tbl:comp_table4}).
 More specifically, ResVGAE outputs higher AUC and AP scores than VGAE, and the difference in score is significant when compared with GAE for CiteSeer and PubMed data sets. For the Cora data set, we see that ResGAE performs better than ResVGAE and the other graph autoencoders without residual modules.

Furthermore, in Table \ref{tbl:comp_table5}, we compare our model with other baseline models on link prediction tasks. 
Our model outperforms Spectral Clustering and DeepWalk algorithms in all three datasets and scores a very similar result to the ARVGE algorithm on the Cora and PubMed datasets.

 Tests on publicly available data sets reveal the importance of architectures with residual modules for exploring the interactions among far nodes for link prediction tasks.
In models without residual modules,  variational graph autoencoder scores higher in the CiteSeer data sets whereas graph autoencoder outperforms graph variational autoencoder in the Cora and PubMed data sets.

\begin{table*}[hbt!]
\centering
\caption{Performance comparison of models with other baseline models. \\}
\label{tbl:comp_table5}
\resizebox{1\linewidth}{!}{%
\begin{tabular}{cccccccc}
\hline Models & \multicolumn{2}{c}{ Cora } & \multicolumn{2}{c}{ Citeseer } & PubMed \\
& AUC & AP & AUC & AP & AUC & AP \\

\hline DW & $83.1 \pm 0.01$ & $85.0 \pm 0.00$ & $80.5 \pm 0.02$ & $83.6 \pm 0.01$ & $84.4 \pm 0.00$ & $84.1 \pm 0.00$ \\
SC & $84.6 \pm 0.01$ & $88.5 \pm 0.00$ & $80.5 \pm 0.01$ & $85.0 \pm 0.01$ & $84.2 \pm 0.02$ & $87.8 \pm 0.01$ \\

ARVGE & ${9 2 . 4} \pm {0 . 0 0 4}$ & $92.6 \pm 0.004$ & ${9 2 . 4} \pm {0 . 0 0 3}$ & ${9 3 . 0} \pm {0 . 0 0 3}$ & $96.5 \pm 0.001$ & $96.8 \pm 0.001$ \\
\hline ResVGAE & ${87.93} \pm {1.90}$ & $90.23 \pm 1.35$ & ${88.7} \pm {1.53}$ & ${84.47} \pm {0.53}$ & $94.5 \pm 0.64$ & $94.8 \pm 0.20$ \\
\end{tabular}
}
\end{table*}

\section{Conclusion}\label{sec:conclusion}

In this paper, we introduce ResVGAE, a variational graph
autoencoder with residual modules to mitigate the over-smoothing phenomenon occurring in graph neural networks with deep architectures. 
Results indicate that the proposed model scores similar average precisions for shallow models and outperforms variational graph
autoencoders without residual modules for link prediction task among nodes with multi-hop separation. 

As a future research direction, we plan to test our models' performance on larger benchmark data sets and focus on other tasks such as graph classification, node classification, and graph clustering. Furthermore, we intend to experiment with the effect of the inclusion of residual modules with different distributions for the variational graph autoencoders. 

\backmatter

\bmhead{Acknowledgments}

This work is supported by The Scientific and Technological Research Council of Turkey (TÜBİTAK) under Project No. 121E378 and by Istanbul Technical University (ITU) Vodafone Future Lab under Project No. ITUVF20180901P04.

\section*{Declarations}

\begin{itemize}
\item Funding: Noted in the acknowledgments section
\item Conflict of interest/Competing interests Not applicable
\item Ethics approval: Not applicable
\item Consent to participate: Not applicable
\item Consent for publication: All authors' consent
\item Availability of data and materials: Upon request
\item Code availability: Available upon acceptance and can be shared with editors at any time
\item Authors' contributions: All authors consent
\end{itemize}

\noindent

\bigskip
 \pagebreak

\bibliography{ref}


\end{document}